\documentclass[11pt]{article}
\usepackage[final]{acl}

\usepackage{times}
\usepackage{latexsym}
\usepackage[T1]{fontenc}     
\usepackage[utf8]{inputenc}
\usepackage{microtype}       
\usepackage{inconsolata}     

\usepackage{booktabs}       
\usepackage{multirow}       
\usepackage{makecell}       
\usepackage{tabularx}       
\usepackage{threeparttable} 
\usepackage{supertabular}   
\usepackage{siunitx}        
\usepackage{adjustbox}      

\usepackage{graphicx}
\usepackage{subcaption}     

\usepackage{algorithm}      
\usepackage{algpseudocode}  

\usepackage{listings}
\usepackage{xcolor}
\usepackage{tcolorbox}
\tcbuselibrary{breakable,skins,listings}

\usepackage{amsmath}
\usepackage{amssymb}
\usepackage{pifont}         
\usepackage{enumitem}

\usepackage[capitalise,nameinlink]{cleveref}

\setlist[itemize]{leftmargin=1.2em,topsep=2pt,itemsep=1pt,parsep=0pt}
\setlist[enumerate]{leftmargin=1.4em,topsep=2pt,itemsep=1pt,parsep=0pt}
\setlist[description]{leftmargin=0pt,topsep=2pt,itemsep=1pt,parsep=0pt,style=nextline}

\definecolor{cpInk}{HTML}{1A1A1A}       
\definecolor{cpRule}{HTML}{9AA0A6}      
\definecolor{cpWash}{HTML}{F4F5F7}      
\definecolor{cpBlue}{HTML}{2B5C8A}      
\definecolor{cpGreen}{HTML}{2E6B4F}     
\definecolor{cpAmber}{HTML}{8A6A1F}     
\definecolor{cpPlum}{HTML}{6B3F7A}      
\definecolor{cpRed}{HTML}{A33A32}       

\newenvironment{fitbox}
  {\begin{adjustbox}{max width=\linewidth,center}}
  {\end{adjustbox}}

\newcommand{\tabfont}{\small}

\newcommand{\best}[1]{\textbf{#1}}

\floatname{algorithm}{Algorithm}
\algrenewcommand\algorithmicrequire{\textbf{Input:}}
\algrenewcommand\algorithmicensure{\textbf{Output:}}
\algrenewcommand{\algorithmiccomment}[1]{\hfill\small\textcolor{cpRule}{\# #1}}
\algnewcommand{\LineComment}[1]{\State\small\textcolor{cpRule}{\# #1}}
\newcommand{\algfont}{\small}

\providecommand{\algorithmname}{Algorithm}

\lstdefinestyle{cppython}{
  language=Python,
  basicstyle=\scriptsize\ttfamily\color{cpInk},
  keywordstyle=\color{cpBlue}\bfseries,
  commentstyle=\color{cpGreen},
  stringstyle=\color{cpPlum},
  identifierstyle=\color{cpInk},
  showstringspaces=false,
  breaklines=true,
  breakatwhitespace=false,
  postbreak=\mbox{\textcolor{cpRule}{$\hookrightarrow$}\space},
  tabsize=4,
  columns=fullflexible,
  keepspaces=true,
  upquote=true,
  literate={-}{-}1 {'}{'}1,       
}
\lstdefinestyle{cpjson}{
  basicstyle=\scriptsize\ttfamily\color{cpInk},
  stringstyle=\color{cpPlum},
  showstringspaces=false,
  breaklines=true,
  columns=fullflexible,
  keepspaces=true,
  upquote=true,
}
\lstdefinestyle{cpshell}{
  language=bash,
  basicstyle=\scriptsize\ttfamily\color{cpInk},
  keywordstyle=\color{cpBlue},
  commentstyle=\color{cpGreen},
  showstringspaces=false,
  breaklines=true,
  columns=fullflexible,
}

\tcbset{cpbox/.style={
  enhanced, breakable,
  boxrule=0.5pt, arc=1.5pt,
  left=5pt, right=5pt, top=4pt, bottom=4pt,
  fonttitle=\small\bfseries, coltitle=cpInk,
  attach boxed title to top left={yshift=-2mm,xshift=4mm},
  boxed title style={boxrule=0.3pt, arc=1pt},
}}

\tcbset{cpcode/.style={cpbox, listing only,
  colback=cpWash, colframe=cpRule,
  boxed title style={colback=cpWash!60!white, colframe=cpRule}}}

\newtcblisting{pycode}[1][Python]{cpcode, listing style=cppython, title={#1}}

\newtcblisting{pycodenum}[1][Python]{cpcode, title={#1},
  listing options={style=cppython, numbers=left,
                   numberstyle=\tiny\color{cpRule}, numbersep=6pt}}

\newtcblisting{pycodeplain}{cpcode, listing style=cppython}

\newtcblisting{shellcode}[1][Shell]{cpcode, listing style=cpshell, title={#1}}

\newtcblisting{jsoncode}[1][JSON]{cpcode, listing style=cpjson, title={#1}}

\tcbset{cpprompt/.style={cpbox,
  fontupper=\scriptsize\ttfamily,
  before upper={\parindent0pt\parskip3pt\raggedright\sloppy}}}

\lstdefinestyle{cpprompttext}{
  basicstyle=\scriptsize\ttfamily\color{cpInk},
  breaklines=true, columns=fullflexible, keepspaces=true, upquote=true,
  showstringspaces=false,
  postbreak=\mbox{\textcolor{cpRule}{$\hookrightarrow$}\space},
}
\tcbset{cppromptverb/.style={cpprompt, listing only, listing style=cpprompttext}}

\newtcblisting{sysprompt}[1][System prompt]{cppromptverb,
  colback=cpWash, colframe=cpRule,
  boxed title style={colback=cpWash!60!white, colframe=cpRule}, title={#1}}

\newtcblisting{userprompt}[1][User prompt]{cppromptverb,
  colback=cpBlue!4!white, colframe=cpBlue!55!white,
  boxed title style={colback=cpBlue!12!white, colframe=cpBlue!55!white}, title={#1}}

\newtcblisting{modelout}[1][Model output]{cppromptverb,
  colback=cpGreen!4!white, colframe=cpGreen!55!white,
  boxed title style={colback=cpGreen!12!white, colframe=cpGreen!55!white}, title={#1}}

\newtcblisting{judgeprompt}[1][Judge prompt]{cppromptverb,
  colback=cpAmber!5!white, colframe=cpAmber!55!white,
  boxed title style={colback=cpAmber!14!white, colframe=cpAmber!55!white}, title={#1}}

\newtcolorbox{syspromptfmt}[1][System prompt]{cpprompt,
  colback=cpWash, colframe=cpRule,
  boxed title style={colback=cpWash!60!white, colframe=cpRule}, title={#1}}
\newtcolorbox{userpromptfmt}[1][User prompt]{cpprompt,
  colback=cpBlue!4!white, colframe=cpBlue!55!white,
  boxed title style={colback=cpBlue!12!white, colframe=cpBlue!55!white}, title={#1}}
\newtcolorbox{modeloutfmt}[1][Model output]{cpprompt,
  colback=cpGreen!4!white, colframe=cpGreen!55!white,
  boxed title style={colback=cpGreen!12!white, colframe=cpGreen!55!white}, title={#1}}

\newtcolorbox{evidence}[1][Retrieved document]{cpbox,
  fontupper=\small, before upper={\parindent0pt\parskip3pt},
  colback=cpPlum!4!white, colframe=cpPlum!55!white,
  boxed title style={colback=cpPlum!12!white, colframe=cpPlum!55!white}, title={#1}}

\newtcolorbox{examplebox}[1][Example]{cpbox,
  fontupper=\small, before upper={\parindent0pt\parskip3pt},
  colback=white, colframe=cpInk!35!white,
  boxed title style={colback=cpInk!8!white, colframe=cpInk!35!white}, title={#1}}

\newif\ifdraftmode
\draftmodetrue            

\makeatletter
\newif\ifanonymous
\ifacl@anonymize\anonymoustrue\else\anonymousfalse\fi
\makeatother

\crefname{algorithm}{Algorithm}{Algorithms}
\Crefname{algorithm}{Algorithm}{Algorithms}
\crefname{section}{Section}{Sections}
\Crefname{section}{Section}{Sections}
\crefname{appsec}{Appendix}{Appendices}
\crefformat{footnote}{#2\footnotemark[#1]#3}

\usepackage{float}  
\newcommand{\lfts}{LabelFusion\hbox{-}TS}
\newcommand{\lf}{LabelFusion}

\title{\lfts{}: Fusing Large Language Models, Transformer Encoders,\\and Financial Time Series for Monetary-Policy Stance Classification}

\author{
    Michael Schlee$^{1}$,
    Fabian Lukassen$^{1}$
    Christoph Weisser$^{2}$\\
    $^{1}$Centre for Statistics, Georg-August-Universit\"at G\"ottingen, Germany\\
    $^{2}$Hochschule Bielefeld (HSBI) - University of Applied Sciences and Arts,
    Bielefeld, Germany\\
    \texttt{michael.schlee@uni-goettingen.de},
    \texttt{fabian.lukassen@uni-goettingen.de},
    \texttt{christoph.weisser@hsbi.de}
}
\begin{document}
\maketitle

\begin{abstract}
Financial text is produced and interpreted within a market environment, yet financial text classifiers almost always receive text alone. We study whether financial time series are useful as an additional input on the task of classifying sentences from Federal Reserve communication as hawkish, dovish, or neutral. Our system, \lfts{}, extends the \lf{} architecture with this modality: a small voting network combines three independently trained components, a fine-tuned RoBERTa encoder, a prompted large language model (LLM), and a fused ensemble of time-series transformers over the market series of the months preceding publication. Because only about a thousand annotated sentences are available for training, the RoBERTa encoder is first pre-trained on sentences annotated automatically by the LLM and only then fine-tuned on the human labels. Trained on Federal Open Market Committee (FOMC) communication up to 2015 and evaluated on 2015--2022, the fused system achieves 70.2\% weighted F1 -- against 64.1\% for the zero-shot LLM -- and overtakes it with as few as 240 human-labelled sentences. We take this as initial evidence for market time series as an input modality in financial text classification.
\end{abstract}

\section{Introduction}

Financial language is produced and interpreted within a quantitative environment. A reader who assesses a central-bank statement, an earnings release, or a news report does so with the recent behaviour of interest rates, inflation, and asset prices in view, and identical wording can support different readings depending on that background. Market data are also unusually convenient as a model input: they are public, machine-readable, available at daily or finer resolution over several decades, and aligned in time with any dated document. Classification models in financial natural language processing, nevertheless, operate almost exclusively on text. Where the two data types are combined, the direction of use is typically the reverse of the one studied here, with textual features added to numerical models to improve forecasts of prices or volatility \citep{sun2026stock, cao2024ecc}; market series as input to text classifiers remain little explored.

\begin{figure}[t]
  \centering
  \includegraphics[width=\columnwidth]{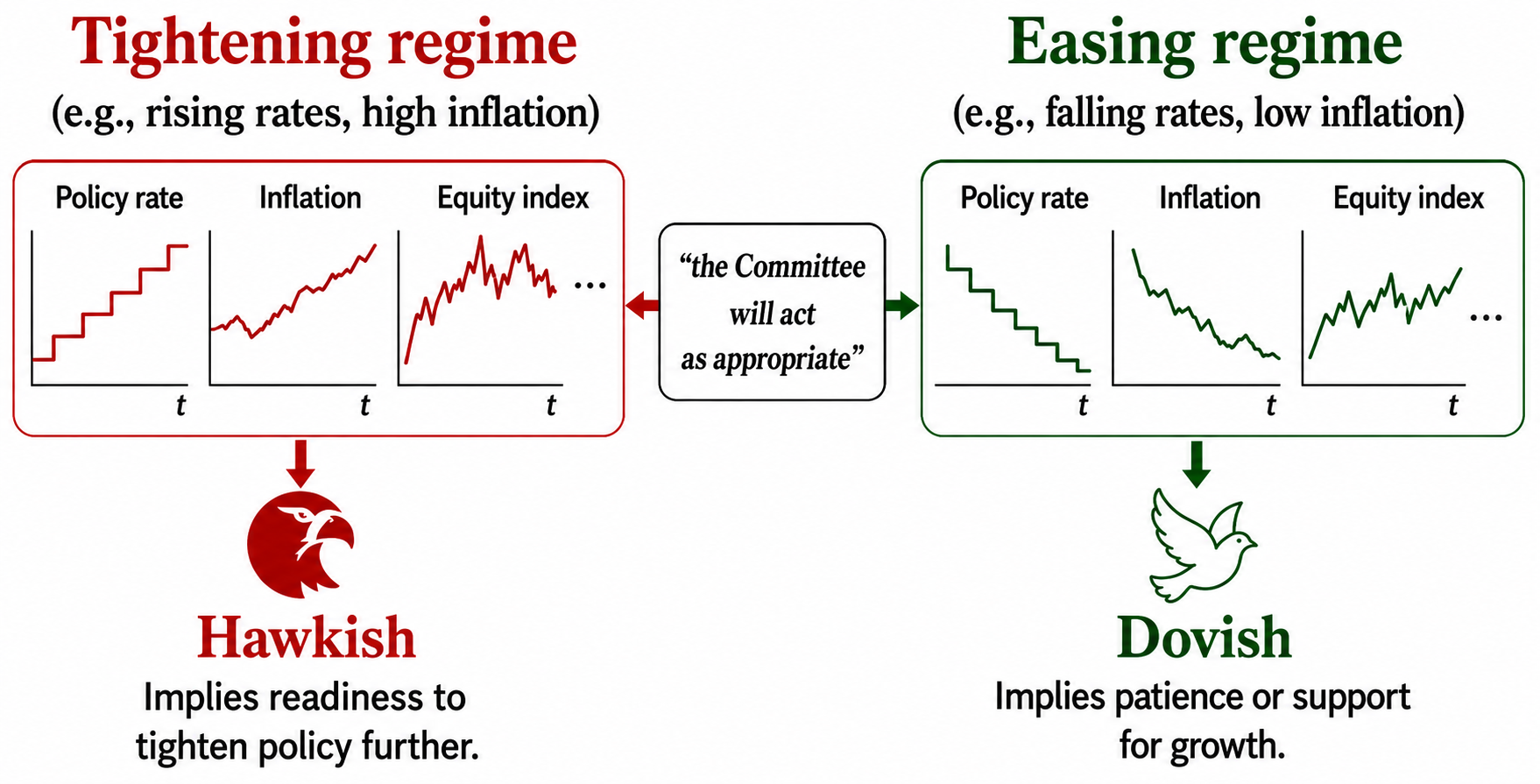}
  \caption{The same sentence under two market regimes. Whether \emph{``the Committee will act as appropriate''} announces further tightening or continued easing is not decided by the words but by the environment they are spoken into; the market series preceding the document carry exactly this information.}
  \label{fig:motivation}
\end{figure}

A step towards using both sources is the \lf{} architecture of \citet{schlee2026labelfusion}, which fuses a prompted large language model with a fine-tuned transformer encoder for financial news classification. The architecture was designed to accommodate additional input sources, and the paper closes by proposing price time series as the next modality. The present work carries out this proposal.

We evaluate it on monetary-policy stance classification \citep{shah2023trillion}, where sentences from Federal Reserve communication are labelled hawkish, dovish, or neutral. The task suits a first test of the modality: central-bank language is deliberately hedged, so that a sentence such as \emph{``the Committee will act as appropriate''} is read differently in a tightening than in an easing cycle (Figure~\ref{fig:motivation}), and every sentence originates from a dated public document, so the market data of the months before it can be attached without further assumptions. Our fused system exceeds a zero-shot large language model when trained on past communication and tested on future statements.

Our contributions are:
\begin{enumerate}[topsep=2pt,itemsep=1pt,parsep=0pt,leftmargin=*]
\item \lfts{}, a recipe for extending intermediate fusion in financial text classification with further input modalities, instantiated here for financial time series.
\item Initial empirical evidence that financial time series can provide context for the classification of financial text across varying data availability regimes.
\end{enumerate}

\section{Related Work}

\paragraph{\lf{}.}
Given a text $x$, \lf{} \citep{schlee2026labelfusion} obtains label scores $z = f_{\mathrm{LLM}}(x)$ from a prompted LLM and a contextual embedding $h = f_{\mathrm{RB}}(x)$ from a fine-tuned RoBERTa encoder and fuses them with a small MLP, $\hat{y} = f_{\mathrm{MLP}}([\,h;z\,])$. Since views are combined as representations rather than as raw features or as separate decisions, this is a hybrid fusion in the taxonomy of \citet{baltrusaitis2017multimodal}. On Reuters-21578, the fusion beats the standalone LLM and RoBERTa classifiers in the full-data regime, while the prompted LLM dominates in the low-data regime. We keep this design unchanged and add one expert.

\paragraph{Monetary-policy stance classification.}
\citet{shah2023trillion} built the benchmark we use: sentences from FOMC meeting minutes, press-conference transcripts, and speeches (1996--2022), labelled hawkish, dovish, or neutral by trained annotators, with baselines up to fine-tuned RoBERTa-large and zero-shot ChatGPT. To our knowledge, no published work uses market data as an \emph{input} to stance classification; the connection between central-bank communication and markets is otherwise well documented \citep{tetlock2007giving,barber2008glitters}.

\paragraph{Time series, text, and distillation.}
Purpose-built series encoders such as PatchTST \citep{nie2023patchtst} treat a time series as a sequence of patches, which suits short fixed windows; the multimodal literature predominantly uses text to help forecast series, whereas we use series to help classify text. Our silver-label stage follows knowledge distillation \citep{hinton2015distilling} in its self-training form \citep{xie2020self}: the LLM acts as a free annotator of in-domain unlabelled text, producing silver pseudo-labels for the student; unlike Xie et al., where injected noise lets the student surpass the teacher, in our setting a subsequent gold fine-tuning stage plays this role."

\section{Data and Task}
\label{sec:data}

\paragraph{Benchmark.}
The benchmark of \citet{shah2023trillion} covers three forms of Federal Reserve communication published between 1996 and 2022: the minutes of FOMC meetings, the transcripts of post-meeting press conferences, and speeches by members of the Board. Trained annotators assigned every sentence the policy direction it signals, hawkish for a tightening of monetary policy, dovish for an easing, and neutral otherwise. The combined release contains 2{,}379 annotated sentences, of which 2{,}312 remain after duplicates and the few sentences with conflicting annotations are removed. Neutral is the largest class (48\%), followed by dovish (27\%) and hawkish (25\%). The task is a single-label three-class classification of one sentence at a time; Appendix~\ref{app:examples} shows examples.

\paragraph{Metric.}
Following the benchmark, we report weighted F1, the class-wise F1 scores averaged with weights proportional to how often each class occurs in the test set, and
\begin{equation}
\mathrm{F1_{w}} \;=\; \sum_{c} \frac{n_c}{N} \cdot \frac{2 P_c R_c}{P_c + R_c},
\label{eq:wf1}
\end{equation}
where $P_c$ and $R_c$ are the precision and recall of class $c$, $n_c$ is the number of test sentences carrying that class and $N$ the size of the test set. The measure therefore rewards precision and recall on every class, but the frequent neutral class dominates the average, unlike the macro average, which weights all three classes equally.

\paragraph{Sentence dates.}
The published sentences state only the year of the document they come from, while both parts of our setup require the exact day of publication: the evaluation splits the data chronologically, and the system pairs every sentence with market data from the period preceding its publication (\S\ref{sec:model}). The documents themselves are published with their dates in the benchmark's repository, which allows the missing date to be recovered by locating the document from which a sentence originates: first by exact match against the sentence lists of the individual documents and for the remaining cases by searching for the normalised sentence in the raw document text. Sentences found in documents of more than one date, mostly formulaic passages that recur across meetings, and sentences found in no document are excluded. The procedure dates 1{,}692 sentences, or 73\% of the benchmark, with a class distribution close to that of the full release; Appendix~\ref{app:dating} gives the details.

\paragraph{Evaluation protocol.}
We split the dated sentences by time: the 1{,}274 sentences up to September 2015 are used for training, with the most recent 15\% held out for model selection, and the 418 sentences from 2015--2022 form the test set. September 2015 serves as the cut-off throughout the paper. The chronological split mirrors deployment, where a model trained on past communication reads future statements, and prevents the leak a random split would allow: sentences from the same period appearing on both sides and revealing the prevailing conditions. A further 96 benchmark sentences occur only in documents published in 2014 or earlier; their exact day is unknown, but they precede the cut-off with certainty, and we use them as additional training sentences for the sentence encoder (\S\ref{sec:model}), the one component that operates on text alone and needs no date.

\paragraph{Market context.}
For each sentence, we build a window of six daily series over the 126 business days ending the day \emph{before} its document date: the federal funds rate, the 2-year Treasury yield, the yield-curve slope, the log equity index, CPI year-over-year inflation, and the unemployment rate, each expressed as its change since the window start. All series are public (FRED). The window ends before the document date, and the macro releases enter with a 45-day publication lag, so it contains only what markets actually knew at the time.

\paragraph{Automatically annotated sentences.}
Expert annotation covers only a fraction of the available text: the training-period documents contain roughly ten times as many sentences as the benchmark annotates. We label 13{,}017 of them with a prompted LLM, using the prompt of \citet{shah2023trillion} (Appendix~\ref{app:prompt}) -- the same model that is also one of the three classifiers our system combines (\S\ref{sec:model}). These \emph{silver labels} are used for a single purpose: pre-training the sentence encoder (\S\ref{sec:model}). The selection excludes benchmark sentences, duplicates, and any sentence with word overlap above $0.85$ with a test sentence, so no test material enters pre-training.

\section{\lfts{}}
\label{sec:model}

\begin{figure}[t]
  \centering
  \includegraphics[width=\columnwidth]{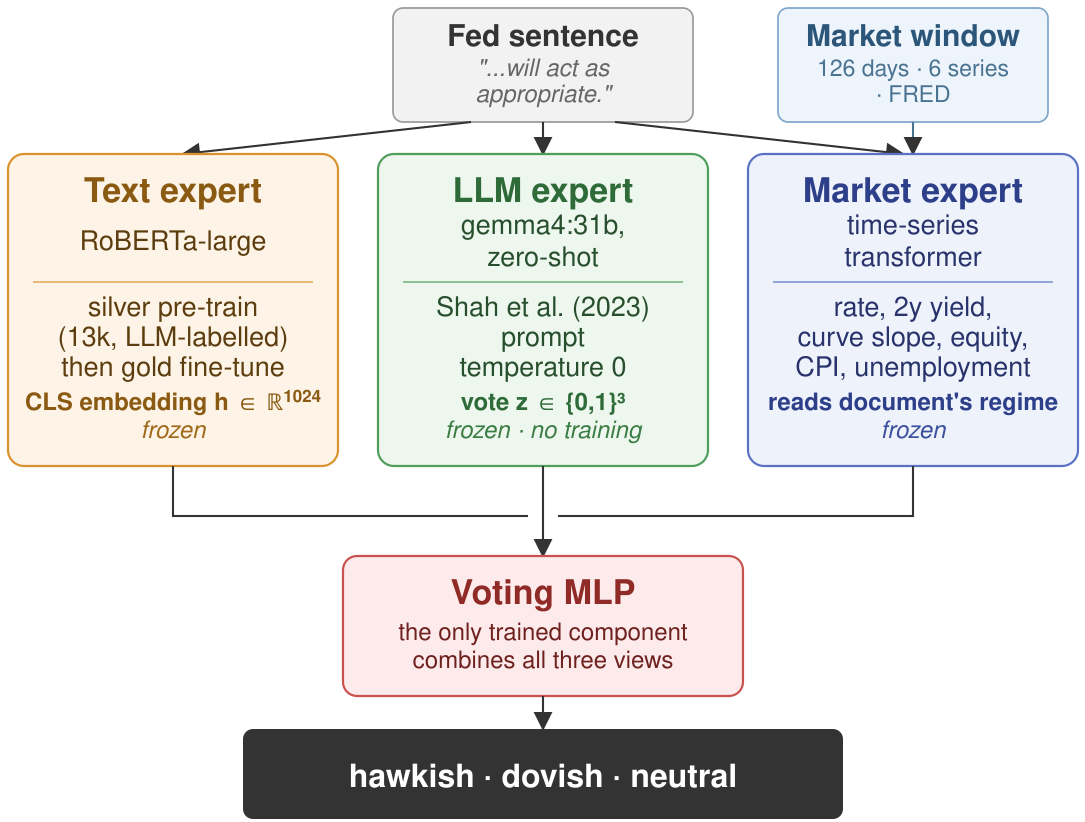}
  \caption{\lfts{}. Three experts are trained independently and frozen: a RoBERTa encoder (pre-trained on LLM-annotated silver sentences, fine-tuned on gold), a prompted LLM voting zero-shot, and a small time-series transformer reading the market window of the sentence's document. A voting MLP combines the three views, exactly as in \lf{} with one additional input.}
  \label{fig:arch}
\end{figure}
\lfts{} keeps the recipe of \lf{}: Three specialised models, called experts in the following, are trained on the task individually, then frozen, and a small voting network is trained separately on their output (Figure~\ref{fig:arch}). Afterwards, we compare LabelFusion-TS with a baseline model, the so-called market expert, which is a standard Transformer encoder \citep{vaswani2017attention} trained from scratch.

\paragraph{Text expert.}
RoBERTa-large \citep{liu2019roberta} is trained in two stages. First, a silver stage: two epochs on the 13k LLM-annotated training-era sentences, which teaches the encoder the task's shape from plentiful but imperfect labels. Then a gold stage: standard fine-tuning on the human-labelled training sentences plus the 96 recovered undated ones. Its CLS embedding $h \in \mathbb{R}^{1024}$ represents the sentence, as in \lf{}.

\paragraph{LLM expert.}
An instruction-tuned open LLM, \texttt{gemma4:31b} at temperature 0, classifies each sentence zero-shot with the classification prompt that \citet{shah2023trillion} used for ChatGPT, reproduced in Appendix~\ref{app:prompt}; its vote is encoded as $z \in \{0,1\}^{3}$. The same model with the same prompt produces the silver labels of \S\ref{sec:data}.

\paragraph{Fusion.}
With all experts frozen, a voting MLP with one hidden layer combines the three views of a sentence,
\begin{equation}
\hat{y} \;=\; f_{\mathrm{MLP}}\big(\big[\,h;\; z;\; m\,\big]\big) \;\in\; [0,1]^{3},
\label{eq:fusion}
\end{equation}
which is \lf{}'s fusion with one added input; removing $m$ recovers \lf{} exactly. Training involves random initialisation at three places, in the text expert, the market expert, and the voting network. To make the results independent of any particular draw, every component is trained with three different random seeds and all $3 \times 3 \times 3 = 27$ combinations are evaluated. We report their mean and, in addition, a \emph{probability ensemble} that averages, per sentence, the class distributions predicted by the 27 systems and can be deployed as a single classifier.

\section{Results}
\label{sec:results}

Table~\ref{tab:main} reports the time-split evaluation. The market expert alone reaches 37.3\%. The text expert reaches 66.1\% and already exceeds the zero-shot LLM (64.1\%) whose silver labels it was pre-trained on. The full fusion reaches 67.4\% on average, with all 27 seed combinations above the LLM, and the probability ensemble of the 27 systems reaches 70.2\%, ahead of every individual component.

\begin{table}[H]
  \centering
  \begin{fitbox}
  \tabfont
  \begin{tabular}{@{}lc@{}}
    \toprule
    & wF1 (\%) \\
    \midrule
    market expert alone                           & 37.3 \\
    LLM zero-shot                                 & 64.1 \\
    RoBERTa-large, silver$\rightarrow$gold        & 66.1 \\
    \midrule
    \lfts{} (ours), mean of 27 runs               & 67.4 \\
    \lfts{} (ours), probability ensemble          & \best{70.2} \\
    \bottomrule
  \end{tabular}
  \end{fitbox}
  \caption{Stance classification on the time-split test set (418 sentences from 2015--2022; models trained on sentences up to 2015). Weighted F1 in percent, following the benchmark's metric; means are over all seed combinations of the trained components. }
  \label{tab:main}
\end{table}

Table~\ref{tab:budget} reports the amount of human labelling required by the system. \lfts{} overtakes zero-shot LLM at roughly 20\% of the training pool ($\approx$ 240 sentences) and remains above it for 100\%, while the LLM row remains constant by construction. \citet{schlee2026labelfusion} report a crossover only after the majority of a substantially larger training set."

\section{Discussion}
\label{sec:discussion}
\paragraph{Why market context can help.}
The market--label connection is visible before any model is trained: grouping training-year sentences by the federal-funds-rate move over the preceding 126 business days, 66\% are hawkish when the rate rose by at least 10 basis points, versus only 38\% when it fell as much. Hedging often leaves the words alone underdetermined; \emph{``act as appropriate''} (Table~\ref{tab:examples}) appears in a dovish 2019 sentence and a neutral 2000 one. A text-only model under a chronological split cannot observe the test period's environment, while the market window supplies it at test time by construction. This also explains why the market expert is weak alone (37.3\%): all sentences of a document share one window, so it contributes to the environment rather than an independent judgement, useful only in combination with the text views. The same logic should extend to other tasks whose label semantics shift with the economic environment, such as sentiment or risk classification across volatility regimes.

\begin{table}[H]
  \centering
  \begin{fitbox}
  \tabfont
  \begin{tabular}{@{}lcccc@{}}
    \toprule
    gold labels & \lfts{} & RoBERTa & Market & LLM \\
    \midrule
    5\%\phantom{0} (59)    & 63.4 & 61.9 & 36.1 & 64.1 \\
    10\% (119)             & 62.8 & 63.9 & 35.6 & 64.1 \\
    20\% (236)             & 66.3 & 66.3 & 38.4 & 64.1 \\
    40\% (471)             & 68.1 & 65.4 & 37.6 & 64.1 \\
    60\% (710)             & 67.8 & 65.1 & 41.0 & 64.1 \\
    80\% (945)             & 69.7 & 67.4 & 38.7 & 64.1 \\
    100\% (1181)           & 70.2 & 66.1 & 37.3 & 64.1 \\
    \bottomrule
  \end{tabular}
  \end{fitbox}
  \caption{Label-budget sweep: weighted F1 (\%) as a function of the number of gold-labelled training sentences. Silver labels and the market expert's rate-cycle pre-training are label-free and constant across rows. \lfts{} is the probability ensemble; RoBERTa and the market expert are means over their three seeds; the zero-shot LLM uses no training data and is constant.}
  \label{tab:budget}
\end{table}

\section{Conclusion \& Limitations}
Financial text classifiers routinely ignore an input every human analyst consults: the market context. Adding it as an additional expert to an existing fusion architecture, trained with silver labels from the fusion's own LLM and evaluated on a train-on-the-past, test-on-the-future protocol, beats a modern zero-shot LLM with as few as ${\sim}240$ human labels, and a simple seed-grid ensemble adds another three points. Components weak alone can be strong together, and the market time series is a cheap, public, time-aligned signal that there is little reason to keep leaving out. The study is limited to one task and benchmark, a seven-year test span dominated by two unusual regimes, a recovered date subset (73\% of sentences), and silver labels drawn from the same LLM used as the zero-shot baseline — a different teacher could shift the picture.
\newpage
\bibliography{refs}

\newpage

\appendix

\section{Benchmark Examples}
\label{app:examples}

\begin{table*}[t]
  \centering
  \small
  \begin{tabular}{@{}llp{0.66\textwidth}@{}}
    \toprule
    Label & Date & Sentence \\
    \midrule
    hawkish & 2005-05-03 & A discernable upcreep was apparent in survey measures of short- and, to a limited extent, long-term inflation expectations over recent months. \\
    hawkish & 2017-12-13 & Many indicated that they expected cyclical pressures associated with a tightening labor market to show through to higher inflation over the medium term. \\
    \addlinespace[2pt]
    dovish  & 2011-08-09 & Some participants noted that additional asset purchases could be used to provide more accommodation by lowering longer-term interest rates. \\
    dovish  & 2019-06-19 & In light of increased uncertainties and muted inflation pressures, we now emphasize that the Committee will closely monitor the implications of incoming information for the economic outlook and will act as appropriate to sustain the expansion with a strong labor market and inflation near its 2 percent objective. \\
    \addlinespace[2pt]
    neutral & 2000-02-02 & As long as the Federal Reserve is required to set and report ranges for money and debt growth, it should update them as appropriate. \\
    neutral & 2004-12-02 & A commonly used analogy takes the U.S. economy to be an automobile, the FOMC to be the driver, and monetary policy actions to be taps on the accelerator or brake. \\
    \bottomrule
  \end{tabular}
  \caption{Annotated sentences of the benchmark, with the document date recovered by the procedure of Appendix~\ref{app:dating}. The last two rows illustrate the difficulty of the task: both are labelled neutral although they discuss policy instruments, while the phrase \emph{``act as appropriate''} appears in a dovish sentence of 2019 and in a neutral one of 2000.}
  \label{tab:examples}
\end{table*}

\section{Dating Procedure}
\label{app:dating}

The benchmark releases sentences with labels and a year, but not the document date needed to attach market data. All source documents are public and are included in the benchmark's repository as date-named files. We match each sentence to its document by exact text match against the per-document sentence files and, where that fails, by containment matching of the normalised sentence against the raw document text; normalisation lowercases and strips all non-alphanumeric characters, so that differences in punctuation, quotation marks, and sentence splitting do not prevent a match. Sentences matching documents on more than one date and sentences matching no document are excluded.

Three cases illustrate the procedure. The sentence \emph{``After precipitous drops in March and April, employment rose strongly in May and June as many people returned to work from temporary layoffs.''} occurs in the sentence file of exactly one document and receives its date, the press conference of 29 July 2020; 1{,}203 sentences are dated by such exact matches. The sentence \emph{``Actual or realized saving depends on the equilibrium values of the real interest rate and other economic variables.''} appears in no sentence file but is found, after normalisation, in the raw text of a single document, a speech of 10 March 2005; containment matching dates 489 further sentences this way. The sentence \emph{``Consistent with its statutory mandate, the Committee seeks to foster maximum employment and price stability.''} is the Committee's standing mandate formula and occurs in 76 documents; no unique date can be assigned, and it is excluded. This yields 1{,}692 uniquely dated sentences of 2{,}312 unique sentences (73\%), spanning January 1996 to September 2022, with label proportions close to the full benchmark (neutral 47\%, dovish 28\%, hawkish 25\%). By source, 893 come from meeting minutes, 489 from speeches, and 310 from press conferences. The year column of the release disagrees with the date of the matched document for the large majority of sentences and is not used. Undated sentences whose matching documents are all dated 2014 or earlier (96 sentences, excluding those with word overlap above $0.85$ with a test sentence) are added to the text expert's training data.

\section{Market Window Details}
\label{app:window}

All series come from FRED, the public database of the Federal Reserve Bank of St.\ Louis, retrieved as CSV files through its unauthenticated download interface (\texttt{fred.stlouisfed.org/graph/fredgraph.csv?id=}\emph{series}); the released code performs the retrieval. Table~\ref{tab:series} lists the six channels and their construction.

\begin{table}[H]
  \centering
  \begin{fitbox}
  \tabfont
  \begin{tabular}{@{}lll@{}}
    \toprule
    Channel & FRED series & Construction \\
    \midrule
    policy rate        & \texttt{DFF}       & level \\
    2-year yield       & \texttt{DGS2}      & level \\
    yield-curve slope  & \texttt{DGS10}, \texttt{DGS2} & difference \\
    equity market      & \texttt{NASDAQCOM} & logarithm of the index \\
    inflation          & \texttt{CPIAUCSL}  & 12-month change, 45-day lag \\
    unemployment       & \texttt{UNRATE}    & level, 45-day lag \\
    \bottomrule
  \end{tabular}
  \end{fitbox}
  \caption{The six market channels. Daily series are used as released; monthly series (inflation, unemployment) are stepped forward to daily frequency and shifted by 45 days to respect their publication delay.}
  \label{tab:series}
\end{table}

A window spans the 126 business days ending the day before the sentence's document date. Within the window, every channel is expressed as its change since the first day of the window, and each channel is standardised with mean and standard deviation computed on the training split only.

\section{LLM Prompt}
\label{app:prompt}

The LLM expert uses, verbatim, the prompt published by \citet{shah2023trillion} for their ChatGPT baseline; \texttt{[sentence]} is replaced by the sentence to classify.

\begin{userprompt}[Classification prompt \citep{shah2023trillion}, verbatim]
Discard all the previous instructions. Behave like you are an expert sentence classifier. Classify the following sentence from FOMC into 'HAWKISH', 'DOVISH', or 'NEUTRAL' class. Label 'HAWKISH' if it is corresponding to tightening of the monetary policy, 'DOVISH' if it is corresponding to easing of the monetary policy, or 'NEUTRAL' if the stance is neutral. Provide the label in the first line and provide a short explanation in the second line. The sentence: [sentence]
\end{userprompt}

The first line of the reply is parsed as the label. The same model and prompt produce the silver labels for the 13{,}017 automatically annotated sentences of \S\ref{sec:data} (99.98\% of replies parse).

\section{Training Details}
\label{app:training}

Table~\ref{tab:hyper} lists the training configuration of the three experts and the voting network; Table~\ref{tab:params} accounts for the market expert's parameters.

\begin{table}[H]
  \centering
  \begin{fitbox}
  \tabfont
  \begin{tabular}{@{}ll@{}}
    \toprule
    \multicolumn{2}{@{}l}{\emph{Text expert}} \\
    encoder            & \texttt{roberta-large}, max.\ 128 tokens \\
    optimiser          & AdamW, learning rate $10^{-5}$ \\
    batching           & size 16, gradient accumulation 2 \\
    loss               & class-weighted cross-entropy \\
    silver stage       & 2 epochs on the 13{,}017 silver sentences \\
    gold stage         & $\le 6$ epochs, epoch selection on validation \\
    precision          & mixed (fp16) \\
    \midrule
    \multicolumn{2}{@{}l}{\emph{Market expert}} \\
    input              & $126 \times 6$ window, 18 patches of 7 days \\
    encoder            & 4 layers, width 64, 8 heads, FF 128 \\
    regularisation     & dropout 0.2 \\
    pre-training       & windows every 2nd business day 1962--2015, \\
                       & \quad rate change in 30 days ($\pm 10$bp, 3 classes) \\
    task training      & Adam, $3 \times 10^{-4}$, $\le 40$ epochs, validation sel. \\
    \midrule
    \multicolumn{2}{@{}l}{\emph{Voting network}} \\
    architecture       & MLP, one hidden layer of 64 \\
    inputs             & concatenated frozen expert outputs \\
    \bottomrule
  \end{tabular}
  \end{fitbox}
  \caption{Training configuration.}
  \label{tab:hyper}
\end{table}

\begin{table}[H]
  \centering
  \begin{fitbox}
  \tabfont
  \begin{tabular}{@{}lr@{}}
    \toprule
    patch projection ($42 \times 64$ + biases)      & 2{,}752 \\
    position embeddings ($18 \times 64$)            & 1{,}152 \\
    encoder layers ($4 \times 33{,}472$)            & 133{,}888 \\
    \quad per layer: attention                      & 16{,}640 \\
    \quad per layer: feed-forward                   & 16{,}576 \\
    \quad per layer: layer norms                    & 256 \\
    output layer norm                               & 128 \\
    classification head                             & 195 \\
    \midrule
    total                                           & 138{,}115 \\
    \bottomrule
  \end{tabular}
  \end{fitbox}
  \caption{Parameter count of the market expert.}
  \label{tab:params}
\end{table}

Every trained component uses seeds $\{1,2,3\}$, giving the 27 systems of \S\ref{sec:model}. Label budgets subsample the gold pool stratified by class; silver data and rate-cycle pre-training stay constant across budgets. The full pipeline, all seeds and all budgets, runs in under two hours on a single T4 GPU; an end-to-end notebook reproducing every number, together with the input bundle and its construction script, will be released (anonymised for review).

\end{document}